\documentclass{article}

\usepackage{arxiv}

\usepackage[utf8]{inputenc} % allow utf-8 input
\usepackage[T1]{fontenc}    % use 8-bit T1 fonts
\usepackage{hyperref}       % hyperlinks
\usepackage{url}            % simple URL typesetting
\usepackage{booktabs}       % professional-quality tables
\usepackage{amsfonts}       % blackboard math symbols
\usepackage{nicefrac}       % compact symbols for 1/2, etc.
\usepackage{microtype}      % microtypography
\usepackage{lipsum}		% Can be removed after putting your text content
\usepackage{graphicx}
\usepackage{amsmath}
\usepackage{array}

\title{A Brief Survey of Associations Between Meta-Learning and General AI}

\date{} 					% Or removing it

\author{ \href{https://orcid.org/0000-0001-7431-1619}{\includegraphics[scale=0.06]{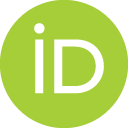}\hspace{1mm} Huimin Peng}\thanks{
		Thank you for all helpful comments! Feel free to leave a message about comments on this manuscript. In case I did not receive email, my personal email is \texttt{974630998@qq.com}. Thanks to github.com/kourgeorge/arxiv-style for this pdf latex template.} \\
	\texttt{peng.huimin.pennie@gmail.com} \\
}

\begin{document}
\maketitle

\begin{abstract}
This paper briefly reviews the history of meta-learning and describes its contribution to general AI. Meta-learning improves model generalization capacity and devises general algorithms applicable to both in-distribution and out-of-distribution tasks potentially. General AI replaces task-specific models with general algorithmic systems introducing higher level of automation in solving diverse tasks using AI. We summarize main contributions of meta-learning to the developments in general AI, including memory module, meta-learner, coevolution, curiosity, forgetting and AI-generating algorithm. We present connections between meta-learning and general AI and discuss how meta-learning can be used to formulate general AI algorithms. 
\end{abstract}

% keywords can be removed
\keywords{Meta-Learning \and General AI \and Coevolution \and Curiosity \and Forgetting}

\section{Introduction}
	\label{introduction}

\subsection{General AI}
\label{generalai}

%\cite{Joshi2019}
Current AI research primarily concentrates attention upon computer vision, natural language processing and automated AI systems, seeking to narrow the performance gap between machine and human beings \cite{Joshi2019}.
% contribution of general AI to the modern world
General AI (AGI, Artificial General Intelligence) seeks algorithms that are as 'smart' as human beings. For example, computer-based systems such as AlphaGo are created and they can defeat even the best human players. General AI aims to design algorithmic models that are applicable to a wide range of tasks. Usually in meta-learning, deep models can only perform self-improvement to solve similar tasks but not vastly different tasks. In AutoML, spaces of deep models are searched through to identify the optimal network for any task in a highly automated fashion. AutoML is a general AI algorithm that can be applied to a wide range of tasks potentially including vastly different ones. Meta-learning pursues speed and accuracy in the self-improvement of deep models to solve unseen tasks. AutoML is slower and needs to accelerate model search through sparse search spaces using meta-learning tools. In the future, meta-learning may develop more general methods that are capable of adapting deep models to solve vastly different tasks more efficiently.

% development of general AI to another realm
Strong AI pursues the development of algorithms comparable or superior to human intelligence and suffers criticism of ethics. General AI belongs to weak AI realm and invents algorithms to handle varied tasks that would require human intelligence to settle \cite{Fjelland2020a}. 
% artificial general intelligence AGI is achievable by the end of this century
There have been ongoing discussions as to whether computer-based algorithms will bring about general AI. As in \cite{Joshi2019}, considering the fast development in AI research and application of AI to almost all areas of human life while exploiting the benefits of big data, general AI will be accomplished in this century.  
%\cite{Fjelland2020a} 
% why general AI will not be realized
In contrast, \cite{Fjelland2020a} believes that computer-based algorithmic models will not realize general AI for the following reasons. First, general AI requires that model account for causalty and perform human-like logic reasoning. Second, there is no explicit representation for tacit knowledge from human comprehension, which is the critical piece of information applied in problem solving. Sometimes human beings are not even aware of the tacit knowledge they manipulate when they complete a mission. To the best of our knowledge, tacit knowledge cannot be represented explicitly in algorithmic general AI models. Third, computer-based learners are not positioned in any real social network. Algorithms do not acquire crucial information externally through versatile interactions within social networks, where subjective and objective information can be balanced to make sensible optimal decisions. 
% what is general AI, belongs to weak AI

% current best general AI algorithms
Current best general AI systems include IBM's \textbf{Watson} and Google's \textbf{AlphaGo}. Watson excels at natural language processing and can be used as AI doctors, which can communicate with patients through dialogues and write prescriptions. Watson has access to history of medical records, related academic literature, previously prescribed medication, etc.
These vital records are thoroughly weighed and processed in Watson to produce recommendations and to assist doctors with routine tasks.  
Furthermore, AlphaGo outperforms human Go players by training supervised deep reinforcement learning models with real games data. 
% list out failure cases of current best AI
Despite the pronounced success of Watson and AlphaGo, even the best general AI systems are susceptible to failure cases. \cite{Fjelland2020a} points out that Watson may not handle vastly different tasks such as unseen out-of-distribution tasks. Furthermore, AlphaGo trained on Go may not be well generalized to other games with different explicit rules. 
% analyze why these current best AI failed in these cases
Although deep learning models succeed in solving trained tasks with high accuracy, in reality, we are faced with more challenging situations where we should generalize to dynamic unseen tasks. 
% possible improvements upon current best AI to avoid failures
Moreover, both causality and tacit knowledge should be considered in algorithmic models to realize general AI \cite{Fjelland2020a}.
% tacit knowledge representation
On one hand, tacit knowledge may be implicitly represented by deep neural network models where hidden layers process a large amount of high-dimensional information. 
% world depiction in the model
On the other hand, although algorithmic models can neither be placed within real social networks nor learn from subjective emotions, these elements may be embedded in algorithms to compensate for over-simplified model specifications.

%\cite{Bellemare2015}
% introduce a platform to test general AI algorithms
ALE (Arcade Learning Environment) \cite{Bellemare2015} is a software platform devised to evaluate the performance of general AI algorithms. In ALE, algorithms are scored using their performances on Atari 2600 games. A game is seen as a task. Algorithms are trained on several games and tested on others to judge generalization capacity.  Game types in the game pool are taken to be as diverse as possible to best measure the generalization capacity of general AI algorithms. 
The best general AI model shows the best performance on the greatest number of games in ALE.
% general AI is supposed to excel in all games (all tasks, all settings)
General AI is supposed to score high under all game modes in all games. 
% proportion of all tasks that can be solved well with each general AI
The proportion of all tasks that can be scored well with the general AI system is its primary performance measure. 
% how well it performs on all tasks
Meta-learning ingredients are integrated into general AI to achieve better model generalization capacity and to accelerate optimal model search in automated machine learning. 

%\cite{Szegedy2020a}
% learn to reason using autoformalization
General AI can be realized by learning to reason automatically from natural language using \textbf{autoformalization} methods \cite{Szegedy2020a}. Autoformalization performs causal reasoning in an automated way through deep learning. It provides an attempted solution to the first hurdle of general AI stressed in \cite{Fjelland2020a}.
% what is autoformalization
Autoformalization system tranforms natural language into uniform representations of mathematical reasoning, from which deductions and inductions can be made using rules of mathematical reasoning. Autoformalization allows general AI algorithms to standardize diverse natural language and to reason mathematically in an automated way.
% mathmatical reasoning is general applicable to all tasks
It is assumed that mathematical reasoning is fundamental  and that autoformalization technique is applicable to all reasoning tasks \cite{Szegedy2020a}. 
% what is autoformalization composed of
Autoformalization is comprised of the following three components: (1) A dimension-reduction model that tranforms standard mathematical reasoning statements into formal feature embedding, (2) A translation model that translates diverse natural language into standard mathematical statements and extracts the corresponding feature embedding, (3) An exploration model that searches for the premise model and the reasoning conversion between premise and current embedding.
% self-improvement upon mathmatical reasoning, generalize to different tasks
Automated autoformalization system can be self-improved to generalize to different reasoning tasks. However, searching for the optimal reasoning conversion using open-ended exploration is still challenging in the general AI system. Meta-learner can be used to accumulate reasoning experiences and to guide the search for reasoning conversion in the exploration model.

\subsection{Meta-Learning}
\label{meta-learning}

%\cite{Peng2020}
% comprehensive survey of recent advances in few-shot meta-learning
A comprehensive survey of recent advances in few-shot meta-learning is provided in \cite{Peng2020}. Updating deep learning models for adaptation to few-shot unseen tasks has attracted significant attention. 
% meta-learning is applicable to few-shot tasks, 
Meta-learning improves model generalization capacity of deep neural networks to achieve the objective of 'learning-to-learn'. 
% both in-distribution tasks and out-of-distribution tasks
For in-distribution tasks (similar to trained tasks), meta-learner identifies model training results of the most similar tasks, 
% for in-distribution similar tasks, use end-to-end direct inference based upon similarity
and directly applies these results to solve in-distribution tasks. 
% for out-of-distribution vastly different tasks, use generalization modules, 
For out-of-distribution tasks (different from trained tasks), meta-learner performs model adaptation using few-shot task data through model generalization modules. Meta-learner aggregates prior model training experiences to confirm the most promising direction for current model adaptation.
% which may cover domain adaptation, mapping from one domain to another
But meta-learning is not constrained in few-shot deep learning and has also a significant influence upon automated general AI. This part was not covered in \cite{Peng2020}. To complement for this part, this manuscript is mainly about the contribution of meta-learning to general AI. 
	
%\cite{Clune2019}	
% AI-GA
AI-GA (\textbf{AI-Generating Algorithm}) \cite{Clune2019} applies meta-learning architectures to construct general AI algorithms . 
% three components in AI-GAs
AI-GA is comprised of three components: (1) A meta-learning architecture that provides the framework of general AI algorithm, (2) A meta-learning algorithm that implements self-improvement for specified machine learning models, (3) An effective generative mechanism that creates inceasingly complicated tasks. For reinforcement learning tasks, a generative scheme offers increasingly complex environments which are used as new tasks to train the solver. 
% how it can be applicable to general settings
A general AI algorithm should be applicable to diverse task settings and should be able to automate any machine learning model. 
% it belongs to general AI concepts
AI-GA framework combines generative algorithms and meta-learning architectures to create a general AI system that can settle a wide variety of tasks. 
% its connection with coevolution algorithm in genetic algorithms
In generative scheme, AI-GA considers task-solver pairs, in which task evolves to be more complex and solver performs self-improvement to solve generated tasks. 
% how it is a meta-learning algorithm
Meta-learner accumulates model training experiences and guides the self-improvement of learner to solve generated tasks efficiently. Meta-learner is the 'brain' that looks back and makes analysis so that future exploration is more sensible.
% development of meta-learning towards AI-GAs, milestones
Meta-learner can automate the self-improvement process of any machine learning algorithm to solve unseen tasks. 
% controlled evolution, open-ended evolution

% AutoML is controlled evolution/evolution
AI-GA guides the self-improvement of solver through meta-learner and task generative mechanism \cite{Clune2019}. AI-GA trains a meta-learning-based general AI system using generated tasks under the guidance of 'upper-level' meta-learner. In the real 'world', task generative mechanism is often governed by hyperparameters which vary cyclically so that generated tasks are not far off the 'mainstream'. 
Since open-ended unbounded task generation may not provide useful tasks which best help learner self-update to overcome the most urgent disadvantage, task generation should be controlled and guided by a meta-learner. 
In addition, careful experimental design may be integrated in the task generative mechanism so that it explores the unknown 'world' more efficiently. 
AI-GA uses meta-learning architectures to construct general AI algorithms. Any machine learning model can be regarded as the base learner / solver. Meta-learner provides automated adaptation of the machine learning model to solve any task. In the generative mechanism, tasks evolve to be increasingly complex and correspondingly the solver is self-updated to solve generated tasks with high performance.

Meta-learning methods are integrated into general AI algorithms to achieve high automation and extensive applicability.
\textbf{AutoML} (Automated Machine Learning) searches for the global optimal neural network for any task in an automated way. In this sense, AutoML is also a general AI algorithm since it automates the process of using neural network model to solve any task and to achieve highest prediction accuracy.
Meta-learning models are integrated within AutoML where meta-learner aggregates training experiences, predicts performance of each hyperparameter combination and points out the most promising hyperparameter combination to explore in the future. Memory module stores long-term model training experiences so that long-range dependence can be considered in the model. Finding a new network that outperforms the current best neural network is regarded as the current task to fulfill. Network model is self-updated to solve the current task and a better neural network model is found. For any task, meta-learner utilizes previous model training experiences to offer a promising initial model, which is not far away from the global best model for this task. 
The aforementioned \textbf{autoformalization} \cite{Szegedy2020a} is a general AI algorithm that conducts automated mathematical reasoning from diverse natural language. 
Meta-learner can be used to accelerate the premise and conversion search in the exploration model of the automated autoformalization reasoning system.  
	
This paper provides a brief review of meta-learning methods that contribute to the realization of general AI and is organized as follows. Section \ref{neuralnetworkdesign} lists out versatile neural network designs in meta-learning that may be applied in general AI algorithms. Section \ref{coevolution} summarizes coevolution frameworks in meta-learning that can be used to construct general AI methods. Section \ref{curiosity} describes artificial curiosity which is a meta-learning concept that can be applied to avoid local optima in general AI optimization. Section \ref{forgetting} surveys forgetting mechanisms in machine learning models. Section \ref{conclusionanddiscussion} summarizes the contribution of these meta-learning frameworks to the developments in general AI.

\section{Neural Network Design}
\label{neuralnetworkdesign}

%\cite{Schmidhuber2015a}
% developments in deep learning
\cite{Schmidhuber2015a} reviews the significant historical breakthroughs in deep learning covering a complete archive of neural network model specifications.
% metalearning, this word is used in this review paper, not meta-learning
%In \cite{Schmidhuber2015a}, area 'metalearning' is equal to 'meta-learning', and sometimes equal to 'meta learning'. 
% meta-learning is used in Chelsea's paper
%In MAML paper, 'meta-learning' is used. They all refer to the same field of meta-learning.
% metalearning does not account for a large proportion in this survey paper
Meta-learning does not constitute a large proportion in this survey paper,
% except that RNN and LSTM are natural metalearning systems to obtain deep learners 
except that RNN and LSTM are both meta-learning systems to adapt deep learners
% efficiently on an unseen task
efficiently to unseen tasks. 
% RNN and LSTM can be very deep, accounting for countless number of layers
RNN and LSTM can continually perform self-improvement through recursive cells. RNN and LSTM can be very deep including more than 1,000 layers, and gradients in backpropagation do not explode or vanish.
% thus can be used for long-range dependence modeling
RNN and LSTM can process long time series data and can be used to model long-range dependence.
Meta-learning models pursue speed and accuracy and can adapt deep learning models to solve unseen few-shot tasks efficiently.
Meta-learning may be exploited as a complement module of deep learning to increase the generalization capacity of deep neural networks and to infuse higher level of automation into deep models.

%\cite{Hochreiter1997} 
% long short-term memory
LSTM can be viewed as a meta-learning system which devises internal memory modules to store model training experiences, recursive cells to perform self-improvement, multiplicative gates to control input and output, and forget gates to discard redundant past information \cite{Hochreiter1997}. 
% self-recursive neural network
LSTM is a complete system for meta-learning and can be used as a base learner or meta-learner in which stochastic gradient descent self-update is performed. 
%\cite{Schmidhuber2015b}
% RNNAI
RNNAI is a meta-learning framework, where both base learner and meta-learner are specified as RNN models \cite{Schmidhuber2015b}. 
% base learner = RNN
% meta-learner = RNN
% parameter update concessively in base learner and meta-learner
In RNNAI, parameters in base learner and meta-learner are updated alternately. Parameter update in base learner depends upon parameter update in the meta-learner. Conversely, parameter update in meta-learner also depends upon parameter update in the base learner. Communication between base learner and meta-learner makes model training more efficient. Either efficient base learner or efficient meta-learner should improve training efficiency of the overall meta-learning system. Usually self-update of base learner is posited to be  more efficient and meta-learner is updated more slowly to aggregate training experiences. 

%\cite{Srivastava2015b}
% highway network
In highway network \cite{Srivastava2015b}, different layers in neural network are connected using highway shortcuts allowing information to flow directly from one layer to another. It is often compared with ResNet which has proven to be very effective with thousands of layers and with wide width. ResNet uses a shortcut with identity-mapping between adjacent layers. Identity mapping is critical to guarantee that gradients in backpropagation do not explode or vanish so that neural network can be very deep contributing to its superior predictive performance. Different from ResNet, shortcuts in highway network are controlled by gates, where gate function can be specified as a linear combination of signals. 
% shortcut, information flow
Highway network may contain shortcuts between distant layers which can model long-range dependence. In AutoML, highway shortcuts may be considered as discrete hyperparameters to be optimized. Highway network notably enriches the flexibility in neural network model specifications, which can increase both level of automation and performance of optimal model. 
% gate function in shortcut
% equivalence of gate and coefficient
Gate functions should be specified in a similar way to the identity mapping used in ResNet. With properly specified gate functions, gradients backpropagation is stable through many layers allowing highway network to be very deep.

\section{Coevolution}
\label{coevolution}

We know that communication between base learner and meta-learner improves the overall training efficiency of the meta-learning system. Similarly, coevolution between related modules improves the overall training efficiency of the general AI system. 

\subsection{Association Between Coevolution and Algorithms}

Coevolution is common in nature where evolution of one creature depends upon the evolution of another different but related creature. In algorithms, evolution stands for self-improvement or self-update of learner. There are several forms of analogies between coevolution and general AI algorithms. First, coevolution utilizes the communication between self-updates of multiple learners to improve the overall training efficiency of all learners. For example, in meta-learning framework, evolution of base learner depends upon the evolution of meta-learner and evolution of meta-learner also depends upon the evolution of base learner. Second, evolution of one algorithm depends upon the evolution of other algorithms. For instance, after discovering a better feature extraction model, embedding models of former algorithms may be updated with the current best in pursuit of better performance. Third, evolution of solver depends upon the evolution of task. As in AI-GA, generated tasks evolve to be more complex and solver is self-updated correspondingly to settle new tasks. Fourth, evolution of future algorithms depends upon evolution of the current best model. For example, AutoML constantly searches for better neural network models to defeat the current best. 

As in \cite{Fjelland2020a}, a qualified general AI model should be placed within the real 'world' where different learners affect each other and objective or subjective information is obtained to make wise decisions. 
Coevolution between algorithms is such a way to place general AI models within 'world'. 
Relationship between coevolved model components may be cooperative or competitive to each other or both. 
A communication mechanism is devised between different algorithmic models and allows the self-improvement processes of different modules to interact with each other. When algorithms affect each other and evolve collectively, overall all algorithms are trained more efficiently. 
Meta-learning has memory modules to store training results. Coevolution mechanisms are used to model the self-improvement of multiple algorithms.
Meta-learning system integrated with a coevolution module can be used to formulate general AI methods. 

Coevolution mechanism in algorithms resembles coevolution in nature. 
%\cite{Burdon1992}
% gene-for-gene (GFG) coevolution, explain its process, in steps
% resistance in the host
% avirulence in the parasite, non-poison parasite, no-harm parasite
% gene for resistance (R)
% gene for avirulence (V)
% resistance is dominant
% avirulence is dominant
% study upon genetic determinants of resistance and avirulence
\cite{Burdon1992} reviews all significant contributions to research on the genetic process in coevolution. \cite{Burdon1992} also studies the genetic determinants of resistance in host and avirulence in parasite. 
% GFG coevolution between host and pathogen
% metapopulation hierarchy affects GFG coevolution
% metapopulation structure + underlying genetic process driving GFG coevolution
%\cite{Anderson1982}
In nature, coevolution processes are driven by selective forces from internal or external sources \cite{Anderson1982}. Individuals with the highest fitness scores are retained within each generation and perceived as seeds for further mutations. 
In AutoML, selective forces can be viewed as the current best model to outperform, and the corresponding fitness score is whether new algorithms outperform the current best model.
In meta-learning framework, selective forces can be viewed as the meta-loss function to minimize, and the corresponding fitness score is the overall prediction accuracy of current individual algorithms.
In algorithms infused with coevolution mechanisms, selective forces and fitness scores depend upon task objectives and can be flexibly specified in general AI methods.

% model underlying genetic process of coevolution between host and parasite
A differential equation model is composed in \cite{Anderson1982} to describe the underlying genetic process of coevolution between host and parasite. Parasites seek to infect hosts and the selective force is that parasites evolve to be unharmful (avirulent) to hosts. 
Empirically, resistance is the dominant feature in hosts and avirulence is the dominant feature in parasites. 
% selection pressure of coevolution: favors empirically dominant features
In nature, selection pressure favors these empirically dominant features,
% fitness of genotype is dependent upon genetic frequencies of resistance or avirulence
and the corresponding fitness scores hinges upon observed genetic frequencies of favored features. 
In algorithms, selection pressure resembles the regularity terms in objective function which alleviates over-fitting and restricts model complexity to be moderate. 
% genetic process driving GFG coevolution depends upon relation between host and parasite
Genetic processes of coevolution rely upon relation between host and parasite,
% which is empirically estimated from metapopulation, using frequencies
which can be empirically estimated from metapopulations using genetic frequencies. %of resistant hosts and avirulent parasites.
% mixture of genetic process driving coevolution
Genetic process behind coevolution is a linear combination of genetic processes for resistant host and avirulent parasite, resistant host and virulent parasite, susceptible host and avirulent parasite, susceptible host and virulent parasite.
% coevolution determines the future relation between host and parasites
Current coevolution process determines the future dominant features.

To see the influence of coevolution concept upon algorithms, we reveal the analogies between natural coevolution and algorithmic models. 
% meta-learner = host
In meta-learning framework, meta-learner can be seen as 'host'
% base learner = parasites
and base learner can be seen as 'parasite'.
% maximize fitness between host and parasites, minimize avirulence of parasites
From coevolution perspective, objective is to maximize fitness score (overall predictive performance) of 'host' and 'parasite' and to maximize avirulence (task-specific performance) of 'parasite'. From algorithmic perspective, objective is to maximize overall predictive performance of 'host' and 'parasite' and to maximize task-specific performance of 'parasite'. 
% learner = host
In AutoML, the current best model can be seen as 'host'
% tester = parasites
and new algorithms to be tested can be seen as 'parasites'.
% maximize resistance of host
From coevolution perspective, objective is to maximize resistance (current best model performance) of 'host' and to maximize avirulence (outperform current best model) of 'parasites'. From algorithmic perspective, objective is to maximize current best model performance of 'host' and to maximize the excess over current best model performance of 'parasites'.
% task = host
In AI-GA, task can be seen as 'host'
% learner = parasite compatible with host
and the corresponding solver can be seen as 'parasite'. 
From coevolution perspective, objective is to maximize avirulence (task-specific performance) of 'parasites'. From algorithmic perspective, objective is to maximize task-specific performance of 'parasites'. 
In AI-GA, task evolution may be specified to be open-ended, bounded in cycles or conditional upon solver evolution. Eventually, complex tasks that cannot be solved by training from scratch can be settled by AI-GA.

%\cite{Anderson1990}
% emergence of new parasites, AIDS, HIV, mRNA 
Emergence of new parasite such as new pathogen can be explained in coevolution models  \cite{Anderson1990}.
% coevolution learns best possible relation between metapopulations
As mentioned earlier, current coevolution model determines the most probable relation between host and parasite in the future.
% how to modify genetic process so that emergence of new parasites can be explained
Whether an unseen pathogen will emerge depends upon the tradeoff between transmission and avirulence of pathogens. Related features include the number of infected hosts, death rate and recovery rate of infected hosts, death rate of hosts subject to other causes, etc. 
% it may take a long time for new solution to a complex task to be found
It may take a long time before a new pathogen emerges. Likewise in general AI models, it may take a long time before a valid solution is reached. Analogous to transmissibility and avirulence of parasites, a valid general AI model should settle diverse tasks (transmissibility) with high performance (avirulence). 

%\cite{Bergelson2001}
% four types of coevolution in nature
Victim-exploiter coevolution in nature is summarized to be in four types, as in \cite{Bergelson2001}. First, \textbf{predator-prey} is a zero-sum game where predator maximizes number of prey captured and prey minimizes its loss to predator. In game theory, more types of game models may be applied to depict relation between predator and prey. In algorithms, both predator and prey are seen as learners. Through game-theory optimization of reward functions, predator and prey learners achieve the best outcome under the constraint from each other. Second, \textbf{host-parasite} is a system where host evolves to be resistant to parasite and parasite evolves to be unharmful to host. Third, \textbf{host-pathogen} is a system where host is resistant to pathogen and pathogen is harmful to host. Fourth, \textbf{plant-herbivore} is a system where plant is prey and herbivore is predator. There are also other forms of coevolution. An example of coevolution in birds is \textbf{brood parasitism} where one species lays eggs in another species' nests \cite{Rothstein1990a}. Disguise of eggs affects the detection rate of fake eggs and may result in rejection of eggs by another species.
% provide algorithm formulas for these four types of coevolution
% coevolution dynamics represented by genetic progression
Coevolution dynamics are represented by genetic processes where 
% hyper-parameter cycles to limit feature growth/ complexication growth
hyperparameters vary in cycles to limit feature size and model complexication growth.
% only limited growth is allowed in nature
Only restricted model complexication growth exists in nature.
% selection pressure, cost of antagonism, cost of attack
In general AI algorithms, hyperparameters may be posited to vary in cycles so that model complexity growth is bounded. 

%\cite{Rothstein1990a}
% brood parasitism for birds
% acceptance of bird eggs
% rejection of bird eggs
% egg color
% provide algorithm formula for this type of coevolution
% hybrid children, possible chance for creation

\subsection{General AI Algorithms Based Upon Coevolution}

Previous section introduces the analogies between general AI algorithms and natural coevolution phenomenon based upon intuitive descriptions. In this section, applications of coevolution in general AI algorithms are summarized in more detail laying out two main frameworks: coevolution between learners, coevolution between task and solver. 
Generally, too many coevolution modules contain too many rounds of performance evaluations leading to lower algorithmic efficiency. 
Too much competition or improper cooperation between groups may harm the self-improvement of learners.  
However, properly specified coevolution modules are encouraged in general AI models.

\subsubsection{Coevolution Between Learners}

%\cite{Monroy2006}
% setup of coevolution using neural network models
Coevolution mechanism in general AI framework may be in two forms. One is the coevolution between machine learning algorithms \cite{Monroy2006}. Another is the coevolution between task and the corresponding learner, as in POWERPLAY \cite{Schmidhuber2013} and AI-GA \cite{Clune2019}. 
Coevolution is not limited to these two forms. This section uses these two coevolution specifications as examples to illustrate the contribution of coevolution to general AI systems. 
Coevolution models the interaction between related components in general AI methods. Coevolution may be posited to be in different explicit forms based upon the concrete task. Coevolution makes use of the mutual information shared between coevolved modules and improves the overall training efficiency of the whole general AI machine. 

\cite{Monroy2006} proposes using coevolution between learners to train neural network models and to achieve general AI. Coevolution evaluates every machine learning algorithm in the 'world' (memory module), which consists of previous learners. Relation between learners is established through the following channels: (1) comparison of learner performance on diverse tasks, (2) interaction effect upon each other's neural model evolution (self-update), etc. For example, objective function of one learner may depend upon performances of cooperators and competitors. Self-improvement of one module may be associated with self-updates of other modules. In coevolution, learners form a 'social network' where they compete, cooperate, communicate, coevolve, etc. This 'social network' of learners improves the training efficiency of all learners overall. 

% definition of learner and tester
In evolutionary multi-objective optimization, evolving neural network models are seen as learners, and testers are trained neural network models used to evaluate learners. The objective is to find a learner that outperforms all testers. A good learner is expected to surpass as many testers as possible. 
In Hall of Fame (HOF) type of memory archive, the best learners from past generations are stored in Coevolutionary Memory (CM) and are used as testers on future learners.
% Layered Pareto Coevolution Archive LAPCA
CM module used in \cite{Monroy2006} is Layered Pareto Coevolution Archive (\textbf{LAPCA}), another type of memory archive which contains not only testers but also learners.
% definition of pareto front
Learners in LAPCA are ordered through Pareto-dominance. 
Learner A Pareto-dominates learner B if learner A outperforms not only all testers defeated by learner B but also other testers. Pareto front consists of the best learners never Pareto-dominated. Under proposed ordering of learners, Pareto front collects all best learners from previous generations. LAPCA contains efficient memory size and conducts efficient performance evaluations of generated learners.

% NeuroEvolution of Augmenting Topologies NEAT
NeuroEvolution of Augmenting Topologies (NEAT) is applied to produce efficient evolutions (self-improvement) of neural network models. In coevolution, neural network model complexity is in monotonic growth and performance evaluations of one model depend upon all other models in the memory archive. 
% how to extract testers from the pool
Testers may be randomly drawn from CM to evaluate performances of existing learners. In LAPCA, learners that never Pareto-dominate others and testers that fail to distinguish any other learners are discarded. Only learners in the top layers under Pareto-dominance ordering are retained in LAPCA. 
% coevolution towards autoML, towards general AI
The general AI algorithm formulated in \cite{Monroy2006} is an evolutionary algorithm combined with a memory archive to store trained learners, such as NEAT with LAPCA.
% fitness function = objective function, in genetic algorithms
After training under this framework, best neural network models are present in LAPCA. This general AI algorithm is applicable to neural network models and other machine learning models. It provides an automated solution to diverse tasks. In conclusion, this general AI framework consists of a self-improvement algorithm and a corresponding memory module, where learners form a 'social network' and perform self-updates collectively through coevolution mechanism.

\subsubsection{Coevolution Between Task and Solver}

%\cite{Schmidhuber2013}
% powerplay
On the other hand, \textbf{POWERPLAY} \cite{Schmidhuber2013} considers the coevolution between task and solver rather than the coevolution between learners \cite{Monroy2006}. 
POWERPLAY concentrates attention upon task-solver pairs, where the solver tackles diverse generated tasks with high performance. Trained solver is a general AI algorithm that can settle different tasks efficiently in an automated way. 
Upon an unseen task, the whole general AI system is self-updated to solve old tasks and the new one. Since the set of unseen tasks is very large, coevolution between task and solver can make the learner to be self-updated too frequent and thus inefficient in solving diverse problems. The threshold for learner self-update may be specified to be higher or adaptive for improved efficiency. 
% consists of three parts: task invention, solver modification, 
% correctness demonstration
POWERPLAY consists of three components: task invention, solver modification, and correctness demonstration. 

% task invention creates simplest task unsolvable by old solver
First, \textbf{task invention module} generates tasks and finds the simplest task that is unsolvable by old solver. Task invention simulates human inner-driven curiosity to pursue challenging tasks and to achieve self-improvement by continuously solving more difficult tasks. Task invention is the most critical component in POWERPLAY since it uses generated tasks to guide the self-improvement of learners. Task invention constitutes a meta-learner which acts as the 'brain' to decide what kinds of further training can best improve the key aspects of current solver. In the end, task invention module identifies failure cases of current solver and offers the simplest failure case as a challenge. In \cite{Monroy2006}, motivation of learner's self-improvement is the external competition or cooperation from other learners in the same 'social network'. Here in POWERPLAY, motivation of learner's self-improvement is the internally self-generated failure cases of current solver. 

% solver modification solves all old tasks and the newly created task
Second, \textbf{solver modification module} takes as input the simplest unsolved task from task invention module, performs solver self-improvement, solves all old tasks and the newly composed failure case. 
% task-solver pair is stored creating a continually growing archive of solutions
Similar to \cite{Monroy2006}, a memory archive and a self-improvement algorithm are equipped in POWERPLAY. But here the continually growing memory archive contains task-solver pairs and the self-improvement algorithm is present in solver modification module.
Finally, \textbf{correctness demonstration module} illustrates that self-updated solver tackles all old tasks and the new challenging task proposed by task invention module with desirable performance. Correctness demonstration can be inefficient since the archive of all old tasks can be very large and evaluations should be performed on all old tasks every time the solver is self-updated. We may avoid this trouble by not requiring the self-improved learner to retain good performance on all old tasks. Instead we revise the general AI system to provide a light-weight solution to every new task encountered in an automated fashion. The light-weight solver only needs to solve the currently generated task well. General AI system is capable of providing such light-weight solvers to diverse tasks. Though computation of a light-weight solver is more efficient, updating the whole general AI system can still be slow. 

% self-reference procedure in powerplay
As a meta-learning framework, POWERPLAY utilizes former model training experiences to accelerate solver self-improvement. Task similarity can be explicitly formulated so that trained models from the most similar tasks can be utilized directly to propose a good initial model. 
% account for external task rather than self-generated tasks
POWERPLAY accounts for both internally self-generated tasks and external tasks, which are considered as challenges to the current solver. Since task generation guides the self-improvement of solver, external tasks represent motivation from external forces that drive the solver improvement. 
%AI-GA similar to POWERPLAY
The concept of AI-GA is analogous to POWERPLAY, where the coevolution between task and solver is considered. The difference is that both coevolution between learners and coevolution between task and solver are utilized in AI-GA. 
%comment coevolution

\section{Curiosity}
\label{curiosity}

Similar to coevolution, curiosity is another concept from meta-learning that can contribute to general AI. Coevolution creates a joint system for associated modules to perform self-improvement collectively allowing learners to communicate, compete, cooperate, etc and allowing learners to coevolve with self-generated tasks. Appropriately specified coevolution improves training efficiency of the whole general AI system. Curiosity can be applied in self-improvement of learners to circumvent local optimas and reach global optima. Curiosity mechanism can be infused within coevolution modules to improve efficiency of learner self-update.

\subsection{Association Between Curiosity and Algorithms}

%\cite{Schmidhuber1999}
% association between curiosity and coevolution
In coevolution-based general AI methods, learner update is influenced by the coevolution between learners or the coevolution between task and solver, which simulates the real world in algorithms and introduces aggregated automation. 
In addition to coevolution, curiosity is another critical contribution from meta-learning realm to general AI.  
Curiosity and coevolution can be implemented together to design versatile general AI algorithms for all types of tasks. 
Curiosity makes the AI algorithm more general by encouraging exploration of more diverse features and more efficient by avoiding over-exploration of close spots \cite{Schmidhuber1999}. 
However, diverging too far away from current exploration spot may not be wise when we are already close to global optimum. At the early stage of exploration, encouraging curiosity can effectively contribute to avoiding local optimum traps. 

% what is artificial curiosity (AC)
Artificial curiosity (AC) can be defined as the unexpectedness of an event \cite{Schmidhuber1999}. For a predictable event, any outcome that is distinct from the most probable outcome is termed as unexpected. Therefore the pre-requisite of unexpectedness is the predictability of an event, the outcome distribution of which can be well estimated. 
Curiosity may be integrated into the objective function where unexpectedness is maximized so that learners explore more diverse features of the search space. For example, 
%\cite{Schmidhuber2010a}
% introduce curiosity/creativity reward in objective function of RL
\cite{Schmidhuber2010a} adds curiosity or creativity reward in the objective function of reinforcement learning so that robots' behaviors are more diverse.  
% what is novel algorithmic predictability
Novel metrics measuring algorithmic predictability may be devised to compute curiosity efficiently in general AI algorithms. 
% how do we discover predictability through coevolution
During the self-improvement of learners, predictability of learner outcome can be examined along with performance evaluations so that unexpectedness in curiosity is measured timely and accurately. 
% define unexpectedness, unpredictable from trained deep model, 
For trained deep neural network, unexpectedness often corresponds to failure cases of deep model. In these failure cases, outcomes cannot be predicted accurately using the trained deep model. 
% wrong prediction from trained deep model
% avoid boredom, avoid tried spots
Curiosity alleviates over-exploration of explored spots and encourages learners to discover new features ignored by other learners. 
% curiosity may turn out to be harmful, criterion to prevent this from happening
Since curiosity may turn out to be harmful sending learners to wander purposelessly in the search region, general AI methods should contain adaptive criteria to control curiosity.

%\cite{Lehman2011}
% evolution with no objective function and with curiosity/novel search alone
Traditionally, evolution is applied to conduct self-improvement of learners. But evolution is often criticized for not being sufficiently efficient. 
Later in pursuit of higher performance, evolution is useful in complex optimization tasks where evolution can alleviate the influence of local optimas. 
Evolution helps skip local optimum and reach global optimum.  
In deep learning, to control over-fitting, regularity terms are included in the objective function so that the deep model does not grow to be too complex. 
Multiple objective functions may be combined in the optimization to balance several objectives simultaneously in the same deep model. 
Here the objective function for evolution (learner self-improvement) may be solely based upon the curiosity criterion, which guides the learner to explore as diverse spots as possible all over the search space \cite{Lehman2011}. 
Performance is evaluated across all learners explored through curiosity mechanism and the best learner is identified.

% motivation for removing objective function
Objective functions can be defined with or without curiosity terms. Traditionally the objective function is defined to be prediction error, reward, etc. 
The benefits of using curiosity-based objective function are as follows. 
First, for different tasks, performance-based objective functions are different but curiosity-based objective functions can be defined in the same way by maximizing the distance between exploration spots. For the same task, the objective function is not unique and may be defined in multiple ways. Curiosity-based objective functions make general AI algorithms applicable to a wider range of tasks. 
Second, most real-life information is obtained and analyzed by human without a clear objective in mind. For example, human sees birds in a forest, recognizes different types of birds and saves information in memory. This process is not driven by any clear reward or objective. Curiosity-based objective functions simulate such unintentional human learning activities which are purely out of curiosity. 
Third, sometimes objective functions can be short-sighted and may misguide learners to be trapped in local optimum \cite{Lehman2011}. For instance, there is a wall intercepted right at the shortest path from start to end. How do learners detour based upon the shortest path motivated by the objective function? In this case, objective functions help trap learners in this local optimum unable to identify a better path. Curiosity-based objective function excels at avoiding local optimum and has a better chance of reaching a desirable solution. 
Fourth, curiosity-based objective forces learners to explore diverse features of search space so that the algorithm has a deeper insight into the overall conditions around the space. Learners accumulates critical knowledge and skills which can be referred to later in other tasks. With curiosity-based objective, memory collects more information and skills which can facilitate model generalization to a wider range of tasks. 
Fifth, similar to POWERPLAY, curiosity-based objective is applied to several self-generated challenging tasks and model training experiences are saved in meta-learner. After long exploration, meta-learner accumulates sufficient amount of model training experiences which can be used to solve more complex tasks. 

% merely reward novelty metric
However, in practice, merely rewarding novelty metrics is not adequate especially in reward-driven cases where maximizing profits is required. In such cases, curiosity-based objective function makes learners wander in the search space and cannot derive an efficient search algorithm for the global optimum. 
% propose a mixture step, alternating between objective-based optimization &
In general AI, performance-based and curiosity-based objectives may be combined to achieve better performance and efficiency. Performance-based and curiosity-based optimization can be combined in the following ways.
% curiosity-based optimization
First, the objective function may be specified as a linear combination of performance-based and curiosity-based terms. In optimization, the objective function is maximized where performance-based and curiosity-based criteria are considered simultaneously to update learners.
Second, a mixture procedure may be applied with steps alternating between performance-based optimization and curiosity-based optimization. 
One performance-based learner update is followed by another curiosity-based learner update. 
Proportion of performance-based steps is adaptive and depends upon the profit-seeking degree of the given task. Curiosity-based steps are primarily at the early stage of exploration and can help avoid local optimum. 
Third, curiosity-based objective acts as a generative model of proposed exploration spots, and performance-based objective is the 'meta-learner' that screens these proposed spots to identify the most promising one. 
Combination of performance-based and curiosity-based objectives may occur in other forms than these three forms. Infusion of curiosity in performance-based optimization can be much more versatile.

Generally, it is not efficient to focus upon either performance or curiosity alone. General AI methods should include both performance-based and curiosity-based ingredients.
For reward-driven tasks, performance-based modules dominate curiosity-based ones. For exploring activities purely out of curiosity, curiosity-based modules dominate performance-based ones.
% association between curiosity & coevolution
In the previous section, we summarize that coevolution is an indispensable component in general AI. Coevolution links related modules and performs self-improvement of these parts collectively to improve overall training efficiency. Curiosity is often applied jointly with coevolution such that self-improvement of learners can avoid local optimum and reach global optimum improving learner performance.
% open-ended evolution leads to growing complexity in model, complexifying algorithm
However, open-ended evolution and unbounded curiosity may adversely affect the training efficiency of general AI algorithms. 
% curiosity objective function, novelty metric
In curiosity-based modules, novelty metrics should be carefully devised to pursue preferred novelty behaviors of specific tasks, which depend upon tacit knowledge obtained through human intervention \cite{Lehman2011}.
Though well-defined novelty metrics may contribute significantly to improved performance, seeking global optimum is still time-consuming especially for large unexplored search space and sparse solutions. 
To our knowledge, if all local optimas are found, then the optima of all local optimas is the global optima. 
The challenge is that learners can identify one or several local optimas, but not all local optimas. 
However, as we find more local optimas, we are closer to the global optima. 
% fitness objective function

\subsection{General AI Algorithms Based Upon Curiosity}

%\subsubsection{Combination of Performance-Based and Curiosity-Based Objectives}
% propose mixture method, conduct novelty-guided search first,
Under meta-learning framework, meta-learner contains a curiosity-based generator of exploration spots. Meta-learner aggregates model training experiences, provides performance-based evaluations of proposed exploration spots, and pinpoints the most promising future exploration spots. 
Meta-learner increases the efficiency of learner self-improvement and reduces the risk of local optimum. 
% then fitness-based optimization using stochastic gradient descent
Performance-based objective is often trained with stochastic gradient descent, and curiosity-based objective is usually applied in evolution or coevolution based self-improvement. 
Therefore, in combination of performance-based and curiosity-based objectives, learner self-udpate hinges upon evolution or coevolution. 
% optima of all local optimas = global optima
It is known that the optima of all local optimas is the global optima. The more local optimas we manage to identify, the closer we are to the global optima. 
% connecting local optimas or peculiar points solves maze problem fast
In the search space, local optimas and peculiar points (boundaries, non-differentiable spots, etc) are potential candidates of the global optima, and they deserve special attention in the search algorithm. 

%\cite{Velez2014}
% in literature, repeated writing on the same concept is required
% so that people understand the concept fully
% identify all peculiar points in search space
% measure their fitness, compare with all local optimas
For exploration spots proposed by the curiosity-based objective, performance-based objective is evaluated directly, or performance prediction using meta-learner is conducted.
Only spots with high performance evaluations or predictions are explored in the future.  
% optima of all local optimas and peculiar points = global optima
% novelty search = meta-learner accumulating fitness of explored points
In novelty-driven search, base learner keeps exploring more spots and meta-learner accumulates performance evaluations of explored spots. 
%				base learner exploring additional points
% divide novelty metric driven search into meta-learner + base learner
% disadvantages of novelty-driven search
% explain in maze problem that novelty-driven search is equivalent to fitness-driven search
In maze problem, performance-based objective is minimization of distance travelled from start to end, and curiosity-based objective is maximization of distance between explored spots and spots to be explored \cite{Velez2014}.
Since both performance-based and curiosity-based criteria are based upon the same distance measure, outcomes from these two criteria are similar. 
Novelty metric in maze tasks may be defined in the following ways: (1) area covered by explored spots \cite{Velez2014}, (2) distance covered by explored spots \cite{Velez2014}, (3) density of explored spots all over the search space \cite{Velez2014}, 
(4) distance between endpoints of explored trajectories \cite{Stanton2016}, (5) distance between explored trajectories \cite{Stanton2016}, etc.
Curiosity-based criteria should be devised carefully in the most appropriate way to improve search efficiency of the global optimum. The most appropriate novelty metric is defined based upon tacit knowledge which relies upon human interpretation of task information and cannot be utilized in an automated way to construct general AI methods. 

There are two types of curiosity-based search: \textbf{novelty search} and \textbf{curiosity search} \cite{Stanton2016}.
% based upon novelty search, propose curiosity search
% novelty search
Novelty search encourages individual learners to explore different spots from others. Novelty search trains learners to solve a particular type of tasks well and to be specialists in an area. 
% curiosity search
Curiosity search encourages each individual to be versatile and to learn as many different exploration skills as possible. 
% comparison between novelty search and curiosity search
Curiosity search trains learners to acquire more types of skills that can be generalized to a wider range of tasks.
Learners trained with curiosity search are more capable and can tackle challenges much more efficiently. 
% disadvantages of novelty search and why it does not help with optimization
Novelty search can be combined with curiosity search to produce learners with maximum number of different skills and maximum behavioral diversity for each skill. 
For example, in maze tasks, novelty search guides learners to explore spots far away and curiosity search rewards learners for acquiring different exploration patterns. 
% components of curiosity search
There are two components in curiosity search \cite{Stanton2016}: intra-life novelty score and fitness function. 
% intra-life novelty score
Intra-life novelty mechanism assigns scores to different exploration patterns of each individual. 
%fitness
Fitness function is an objective that maximizes the diversity of individual exploration patterns. 
Exploration patterns depend upon the following aspects: (1) performance-based objectives, (2) curiosity-based objectives with several novelty metrics, (3) coping with local traps or mechanical emergencies, etc. 
For instance, in maze tasks, exploration patterns contain opening different types of doors, grabbing objects, avoiding stepping on stones, walking, running, jumping, etc. 

Performance-based objectives can be integrated into novelty search and curiosity search so that trained learners acquire diverse skills and perform well at the same time. 
% intra-life novelty compass, how this addresses the disadvantages of novelty-driven search
Intra-life novelty identifies behaviors that can substantially improve understanding of tacit knowledge within the specific task and increase the survival probability of learners. 
Searches purely relying upon curiosity seek to accumulate more skill sets in memory, and often ignore performance evaluations. 
Curiosity-based search is exploratory, seeks self-generated or external challenges, intends to build more capable and more general learners. 
% combination of novelty search and curiosity search, which is 
% combination of novelty-driven search and fitness-driven search
% comparison between purely novelty-driven, curiosity-driven, combo
% novelty metric definitions: 1 distance between endpoints 2 distance between trajectories
% deep model adaptation using novelty-driven search or curiosity search
With novelty search, curiosity search and performance-based evaluations on diverse tasks, general AI system can adapt deep models to solve vastly different tasks. 
Since trained deep learners already possess multiple necessary skills to solve a wide range of tasks, model adaptation will be fast and efficient. 

%\cite{Kawaguchi2017}
% connection between generalization capability and optimization search technique
In learner optimization, performance-based and curiosity-based objectives can be combined to achieve the best search efficiency and to avoid local optimas. 
Model generalization capability is closely associated with learner optimization technique. 
In practice, generalization capabilities of curiosity-based approaches are higher. The reasons are as follows.
Under performance-based objective, model generalization should also be conducted using performance-based methods.
Under curiosity-based objective, model generalization should be executed using curiosity-based adaptation algorithms. 
Since curiosity-based approaches accumulate more model training experiences including both exploration patterns and diverse behaviors, it is probable that curiosity-based model adaptation can solve more diverse tasks including vastly different ones. 

% connection between generalization and neural network complexity measures
On the other hand, model generalization capacity is related to complexity measures of deep neural network models. For deep models, \cite{Kawaguchi2017} summarizes the upper bound of generalization capacity based upon model complexity. As neural network complexity increases, the upper bound of model generalization capacity also rises. 
% complexity measures: several types, comparison, discussion
Neural network complexity is positively associated with the number of weight parameters and the number of layers. 
When we expect general AI systems to solve vastly different tasks, we may choose deeper and wider neural network models. 
% generalization capacity definition, its upper bounds
It is pointed out that the generalization capacity of unsupervised learning models using unlabelled data will generally be higher than supervised learning models \cite{Kawaguchi2017}. Similarly, the generalization capacity of self-supervised learning models using unlabelled data will also be higher overall. 
% generalization capacity versus model with random labels and true labels
Moreover, the generalization capacity of learners using randomly generated data labels will also be higher than learners using true labels. 
% connection between generalization capacity and network weights norms
Taking advantage of all these phenomena,
we may continually improve the generalization capacity of a deep model until it approaches the generality of a general AI method and eventually becomes a general AI algorithm.

\section{Forgetting}
\label{forgetting}

We have summarized coevolution and curiosity which contribute to the construction of general AI methods. This section concerns forgetting which is also an indispensable component in general AI. 

% why do we forget, redundant information, keep core
Why do we need forgetting mechanisms? 
First, meta-learner continually gathers more critical training experiences and forgetting helps avoid memory explosion. Forgetting discards redundant and obsolete memory items and keeps memory size efficient. With forgetting, we concentrate our attention upon the most relevant and critical pieces of information in memory. 
Second, forgetting is necessary especially for certain reward-driven optimization techniques. 
In some reward-driven algorithms, every step has a separate goal of maximizing profits received at that step. But several steps combined may not have maximized the total profits received. Forgetting reverts actions at previous steps to increase combined profits. 
Third, forgetting mechanisms remove training experiences at exploration spots with high regret or high boredom from memory. Regret of any spot is the distance between any exploration spot and the global optima. When regret is high, these exploration spots are far away from global optima and forgetting mechanisms may discard the corresponding training experiences. Boredom of any spot is the similarity between this spot and spots already explored. When boredom is high, meta-learner has already accumulated too many similar training experiences at similar spots and exploring this spot is no longer intriguing. 
% examples where forgetting contribute to better update

%\cite{Ficici2003}
% coevolutionary forgetting 
In HOF or LAPCA type of memory archives within coevolution modules, forgetting mechanisms remove obsolete or dominated learners to maintain high-level competition between learners \cite{Ficici2003}. 
% memory module in meta-learning, forgetting mechanism to avoid memory explosion
In meta-learning, forgetting mechanisms are installed in memory modules to avoid memory explosion and to improve search efficiency. 
% adaptive, meta-learner, forecast which part of memory may never be used
Forgetting mechanisms adaptively remove memory items by forecasting which components in memory have the lowest probability of ever being used again. 
% age, forget oldest memory 
Forgetting mechanisms can also discard the oldest memory when long-range dependence is not considered in the model.  
% redundancy, too many similar memory items, redundant
When too many similar items are saved, forgetting mechanisms can delete redundant information without harming the representativeness of memory. 
%avoid forgetting important information
But we should be careful that forgetting mechanisms would not under-estimate any important information. 
%forget redundant information
%forgetting is dimension reduction, feature extraction, information distillation
%forgetting in memory module is information distillation%efficient memory
Forgetting mechanisms perform feature extraction and information distillation on memory items. 
%save memory size, faster to search for critical information from memory
Saving memory size makes it faster to search for critical information from memory. 
%forgetting in dimension reduction technique is feature extraction%efficient feature size
Forgetting mechanisms can also use dimension reduction techniques to make feature size more efficient. 
% indispensable component in general AI
Forgetting mechanisms are indispensable components in creating general AI algorithms. 

%\cite{VanderWesthuizen2018a}
% lstm forget gate very effective
% forget gate helps with lstm's model performance
Forget gates in LSTM are very effective and are major contributions to its superior model performance \cite{VanderWesthuizen2018a}. 
% curiosity is another form of forgetting by minimizing effect of previous behavior upon current behavior.
Curiosity can be seen as another form of forgetting since it minimizes (forgets) the influence of prior exploration spots upon current spots.
% memory module is commonly used in meta-learning to store critical model training
% experiences.
%\cite{b5}
% difference between lars and lasso solution path
Lars and Lasso are both solutions to optimization of the objective function with L1 regularization.
% lasso includes forgetting mechanism
Lasso includes a forgetting mechanism which deletes previously added feature variables for improved model performance. 
% lars has no forgetting mechanism
But Lars has no forgetting mechanism and keeps all added features in the selected model.
% lars + forgetting/regret mechanism = lasso solution
% lasso solution is solution to optimization problem in penalized regression
% forgetting criterion in lasso
% forgetting = less than algorithmic global optimality
% in some datasets, lars is better than lasso
In practice, Lars is better than Lasso in some datasets and Lasso can also be better than Lars in others. From this perspective, including forgetting mechanisms in algorithms may not be better necessarily. 
% in some datasets, lasso is better than lars
%forgetting versus global optimization
Theoretically forgetting mechanisms bring learners closer to global optima by showing regret on former actions which have not turned out to be beneficial. 
%forget previously accumulated information = regret
%go back to where I was and re-judge previous decision based upon future information, short-sighted quick profits seeking algorithms never reach global optimum
By re-visiting previous decisions, future information is applied to re-judge former actions and to correct sub-optimal choices.

%why? 
%knowledge -> representation -> decomposition -> steps
Decomposition of knowledge representation can be used to construct concrete steps in specific tasks.
%global objective -> decomposition -> sub-goals -> steps
Correspondingly decomposition of global objective can be applied to formulate sub-goals in each concrete step.  
%how does achieving sub-goals contribute to global objective
At each step, sub-parts of knowledge are exploited to realize sub-goals through optimization. 
%how does global objective optimization contribute to global optimization
Under appropriate decomposition scheme, fulfilling all sub-goals should reach the global optima of the final objective. 
%avoid local optimum
However, in practice, task decomposition may not provide steps that lead to global optima.
The reasons are as follows. For computation efficiency, task decomposition may be approximated with simpler mechanisms.  
Sub-parts of knowledge at each step may be insufficient, biased or misleading.
Coevolution modules within decomposed steps may be improperly devised such that computation error at each step may accumulate and explode making algorithms unstable. 
%short-sighted quick profits seeking algorithms versus local optimum
Following decomposed steps may reach other optimas that are approximately close to global optima.
%but connecting all local optimums found will lead to global optimum
Embedding forgetting mechanism allows re-visiting previous steps after future performance is observed, and making convenient adjustments for better ultimate performance.  
%combining several short-sighted quick profits seeking algorithms = global optimum
Apart from forgetting mechanisms, several decomposition schemes may be combined and optimization at each step may communicate with each other for improved performance.

\section{Conclusion and Discussion}
\label{conclusionanddiscussion}

% meta-learning > few-shot meta-learning
This paper briefly summarizes connections between meta-learning and general AI, covering concepts such as coevolution, curiosity and forgetting. Meta-learning is applicable not only to few-shot deep learning, but also to general AI. 
% all these components in meta-learning lead to general AI
Generally, there are many existing paths to realize general AI and meta-learning makes important contributions by supplying useful tools such as memory module, meta-learner, coevolution, curiosity, forgetting, etc.
% general AI can be realized
Though algorithms are not as smart as human intelligence, at least general AI methods strive to approach human wisdom. Hopefully general AI can be realized bringing automation into solving simple, time-consuming, dangerous, laborious or repeated tasks. 

% combination of neural network design (structure), coevolution (training protocol),
In general, flexible neural network designs offer diverse options to process information flow and evolutionary algorithms realize stable efficient self-improvement of learners. 
% curiosity (optimization search), forgetting (tentative may improve generalization)
Curiosity search encourages learners to avoid local optima and to acquire diverse skill sets. Forgetting improves the efficiency of memory module and corrects previously made sub-optimal decisions. 
% how these components deal with tacit knowledge
Coevolution positions self-improvement of learners in a 'social network' of learners and considers competition, cooperation, communication, etc among learners.  
Curiosity metrics, objectives and forgetting mechanisms are often best determined using tacit knowledge. Human interpretation is required to build tacit knowledge within specific tasks into general AI models.
% how these components represent 'world'
The real world is sophisticated and impossible to simulate. General AI algorithms cannot be close to human performance in all kinds of tasks. 
But hopefully with tacit knowledge built into algorithms, general AI can at least solve a wide range of tasks in an automated way.  

%\section*{Acknowledgment}

\bibliographystyle{unsrt}

%\bibliography{refs9}


\begin{thebibliography}{10}
	
	\bibitem{Joshi2019}
	Naveen Joshi.
	\newblock {How Far Are We From Achieving Artificial General Intelligence?},
	2019.
	
	\bibitem{Fjelland2020a}
	Ragnar Fjelland.
	\newblock {Why General Artificial Intelligence will not be Realized}.
	\newblock {\em Humanities and Social Sciences Communications}, 7(1):1--9, 2020.
	
	\bibitem{Bellemare2015}
	Marc~G. Bellemare, Yavar Naddaf, Joel Veness, and Michael Bowling.
	\newblock {The Arcade Learning Environment: An Evaluation Platform for General
		Agents}.
	\newblock {\em IJCAI International Joint Conference on Artificial
		Intelligence}, 2015-January:4148--4152, 2015.
	
	\bibitem{Szegedy2020a}
	Christian Szegedy.
	\newblock {\em {A Promising Path towards Autoformalization and General
			Artificial Intelligence}}, volume 12236 LNAI.
	\newblock Springer International Publishing, 2020.
	
	\bibitem{Peng2020}
	Huimin Peng.
	\newblock {A Comprehensive Overview and Survey of Recent Advances in
		Meta-Learning}.
	\newblock 2020.
	
	\bibitem{Clune2019}
	Jeff Clune.
	\newblock {AI-GAs: AI-Generating Algorithms, an Alternate Paradigm for
		Producing General Artificial Intelligence}.
	\newblock {\em arXiv}, 2019.
	
	\bibitem{Schmidhuber2015a}
	J{\"{u}}rgen Schmidhuber.
	\newblock {Deep Learning in Neural Networks: An Overview}.
	\newblock {\em Neural Networks}, 61:85--117, 2015.
	
	\bibitem{Hochreiter1997}
	Sepp Hochreiter and J{\"{u}}rgen Schmidhuber.
	\newblock {Long Short-Term Memory}.
	\newblock {\em Neural Computation}, 9(8):1735--1780, nov 1997.
	
	\bibitem{Schmidhuber2015b}
	J{\"{u}}rgen Schmidhuber.
	\newblock {On Learning to Think: Algorithmic Information Theory for Novel
		Combinations of Reinforcement Learning Controllers and Recurrent Neural World
		Models}.
	\newblock pages 1--36, 2015.
	
	\bibitem{Srivastava2015b}
	Rupesh~Kumar Srivastava, Klaus Greff, and J{\"{u}}rgen Schmidhuber.
	\newblock {Highway Networks}.
	\newblock 2015.
	
	\bibitem{Burdon1992}
	J.J. Burdon and J.N. Thompson.
	\newblock {Gene-for-Gene Coevolution between Plants and Parasites}.
	\newblock {\em Nature}, 360:121--125, 1992.
	
	\bibitem{Anderson1982}
	R.~M. Anderson and R.~M. May.
	\newblock {Coevolution of Hosts and Parasites}.
	\newblock {\em Parasitology}, 85(2):411--426, 1982.
	
	\bibitem{Anderson1990}
	R.~M. Anderson.
	\newblock {Parasite–Host Coevolution}.
	\newblock {\em Parasitology}, 100(S1):S89--S101, 1990.
	
	\bibitem{Bergelson2001}
	J.~Bergelson, G.~Dwyer, and J.~J. Emerson.
	\newblock {Models and Data on Plant-Enemy Coevolution}.
	\newblock {\em Annual Review of Genetics}, 35:469--499, 2001.
	
	\bibitem{Rothstein1990a}
	Stephen~I. Rothstein.
	\newblock {A Model System for Coevolution: Avian Brood Parasitism}.
	\newblock {\em Annual Review of Ecology and Systematics}, 21(1):481--508, 1990.
	
	\bibitem{Monroy2006}
	German~A. Monroy, Kenneth~O. Stanley, and Risto Miikkulainen.
	\newblock {Coevolution of Neural Networks using a Layered Pareto Archive}.
	\newblock {\em GECCO 2006 - Genetic and Evolutionary Computation Conference},
	1:329--336, 2006.
	
	\bibitem{Schmidhuber2013}
	J{\"{u}}rgen Schmidhuber.
	\newblock {PowerPlay: Training an Increasingly General Problem Solver by
		Continually Searching for the Simplest Still Unsolvable Problem}.
	\newblock {\em Frontiers in Psychology}, 4:1--21, 2013.
	
	\bibitem{Schmidhuber1999}
	J{\"{u}}rgen Schmidhuber.
	\newblock {Artificial Curiosity based on Discovering Novel Algorithmic
		Predictability through Coevolution}.
	\newblock {\em Proceedings of the 1999 Congress on Evolutionary Computation,
		CEC 1999}, 3:1612--1618, 1999.
	
	\bibitem{Schmidhuber2010a}
	J{\"{u}}rgen Schmidhuber.
	\newblock {Formal Theory of Creativity, Fun, and Intrinsic Motivation
		(1990-2010)}.
	\newblock {\em IEEE Transactions on Autonomous Mental Development},
	2(3):230--247, 2010.
	
	\bibitem{Lehman2011}
	Joel Lehman and Kenneth~O. Stanley.
	\newblock {Abandoning Objectives: Evolution Through the Search for Novelty
		Alone}.
	\newblock {\em Evolutionary Computation}, 19(2):189--222, 2011.
	
	\bibitem{Velez2014}
	Roby Velez and Jeff Clune.
	\newblock {Novelty Search Creates Robots with General Skills for Exploration}.
	\newblock {\em GECCO 2014 - Proceedings of the 2014 Genetic and Evolutionary
		Computation Conference}, pages 737--744, 2014.
	
	\bibitem{Stanton2016}
	Christopher Stanton and Jeff Clune.
	\newblock {Curiosity Search: Producing Generalists by Encouraging Individuals
		to Continually Explore and Acquire Skills Throughout Their Lifetime}.
	\newblock {\em PLoS ONE}, 11(9):1--20, 2016.
	
	\bibitem{Kawaguchi2017}
	Behnam Neyshabur, Srinadh Bhojanapalli, David McAllester, and Nathan Srebro.
	\newblock {Exploring Generalization in Deep Learning}.
	\newblock {\em 31st Conference on Neural Information Processing Systems}, 2017.
	
	\bibitem{Ficici2003}
	Sevan~G Ficici and Jordan~B Pollack.
	\newblock {A Game-Theoretic Memory Mechanism for Coevolution}.
	\newblock {\em Proceedings of the 2003 Genetic and Evolutionary Computation
		Conference}, pages 286--297, 2003.
	
	\bibitem{VanderWesthuizen2018a}
	Jos van~der Westhuizen and Joan Lasenby.
	\newblock {The Unreasonable Effectiveness of the Forget Gate}.
	\newblock pages 1--15, 2018.
	
	\bibitem{b5}
	B~Efron, T~Hastie, I~Johnstone, and R~Tibshirani.
	\newblock {Least Angle Regression}.
	\newblock {\em The Annals of Statistics}, 32(2):407--499, 2004.
	
\end{thebibliography}

\end{document}